\documentclass[letterpaper, 10 pt, conference]{ieeeconf}  

\IEEEoverridecommandlockouts                              

\usepackage{booktabs}

\usepackage{graphics} 
\usepackage{graphicx} 
\usepackage{amsmath} 
\usepackage{amssymb}  
\usepackage{xcolor}
\usepackage{lipsum}
\usepackage{soul}
\usepackage{hyperref}
\usepackage{subcaption}
\usepackage[ruled,vlined,linesnumbered,noend]{algorithm2e}
\usepackage{comment}
\usepackage{tabularx}
\usepackage{siunitx}
\usepackage{wrapfig}
\usepackage{multirow}
\usepackage{multicol}
\usepackage{url}
\usepackage{comment}
\usepackage{dirtytalk}
\usepackage{arydshln}
\usepackage{url}

\usepackage[noadjust,compress]{cite}

\title{\LARGE \bf Integrating Model-based Control and RL\\  for Sim2Real Transfer of Tight Insertion Policies}

\author{Isidoros Marougkas$^{*, 1}$, Dhruv Metha Ramesh$^{*, 1}$, Joe H. Doerr$^1$, Edgar Granados$^1$, \\ Aravind Sivaramakrishnan$^2$, Abdeslam Boularias$^1$, and Kostas E. Bekris$^3$
\thanks{$^{*}$The first two authors contributed equally to this paper. $^{1}$Dept. of Computer Science, Rutgers University, NJ, USA. $^{2}$A.S. is affiliated with Amazon.com Inc. $^{3}$K. E. Bekris holds concurrent appointments as a Professor at Rutgers University and as an Amazon Scholar. This paper describes work performed at Rutgers and is not associated with Amazon. The work has been partially supported by NSF awards NRT-FW-HTF 2021628, FRR 2309866 and POSE 2346069. Opinions expressed here are those of the authors and do not reflect positions of the funding agency.  Corresponding author e-mails: {\tt \{im316,ab1544,kostas.bekris\}@cs.rutgers.edu}.}
}

\begin{document}
\maketitle
\thispagestyle{empty}
\pagestyle{empty}

\begin{abstract}

Object insertion under tight tolerances ($< \hspace{-.02in} 1mm$) is an important but challenging assembly task as even small errors can result in undesirable contacts. Recent efforts focused on Reinforcement Learning (RL), which often depends on careful definition of dense reward functions. This work proposes an effective strategy for such tasks that integrates traditional model-based control with RL to achieve improved insertion accuracy. The policy is trained exclusively in simulation and is zero-shot transferred to the real system. It employs a potential field-based controller to acquire a model-based policy for inserting a plug into a socket given full observability in simulation. This policy is then integrated with residual RL, which is trained in simulation given only a sparse, goal-reaching reward. A curriculum scheme over observation noise and action magnitude is used for training the residual RL policy. Both policy components use as input the {\tt SE(3)} poses of both the plug and the socket and return  the plug's {\tt SE(3)} pose transform, which is executed by a robotic arm using a controller.  The integrated policy is deployed on the real system without further training or fine-tuning, given a visual {\tt SE(3)} object tracker. The proposed solution and alternatives are evaluated across a variety of objects and conditions in simulation and reality. The proposed approach outperforms recent RL-based methods in this domain and prior efforts with hybrid policies. Ablations highlight the impact of each component of the approach. For more information please refer to the corresponding
\href{https://dhruvmetha.github.io/insert}{website}.

\end{abstract}

\section{Introduction}
\label{sec:introduction}

This paper addresses object insertion under tight tolerances ($< \hspace{-.02in} 1mm$). Given visual tracking of the {\tt SE(3)} object pose, this work proposes a strategy for learning a policy for tight insertion into a socket. A key feature of the proposed strategy is that it first defines a model-based control solution, which is then complemented with a residual policy trained via Reinforcement Learning (RL) in simulation to address the uncertainty arising from perception noise and contact dynamics. The policy trained in simulation is directly deployable on the real system without any fine-tuning.


{\bf Tight object insertion} is applicable both in industrial and domestic setups, from product part assembly to plugging sockets of home devices. Thus, peg-in-hole challenges have long been the focus of robotics research \cite{lozano1984automatic, suarez2016framework, morgan2021vision,HaonanIROS2024B} as a contact-rich manipulation task. Nevertheless, the sub-millimeter precision required to complete such tasks and the uncertainty regarding the objects' states has limited the real-world deployment of developed solutions.

{\bf Model-based efforts} \cite{morgan2021vision,  wang2019robotic, morgan2023towards, Wu2024Tree} have engineered control policies for insertion that can be effective for a well-instrumented workspace setup. Such solutions, however, are brittle to changes in the workspace, and do not generalize easily to new objects. Recent {\bf data-driven approaches} have attempted to solve the problem by learning policies either from human demonstrations \cite{wen2022demonstrate,nair2023learning} or from online interaction via RL \cite{tang2023industreal,Luo2024SERLAS,Kulkarni2022AIS}. They can generalize to new objects, nevertheless, they require significant demonstration effort, reward engineering, and incur high sample complexity. 
In particular, recent work that is closely related to this paper, IndustReal \cite{tang2023industreal},
demonstrated peg insertion accuracy at  $\sim \hspace{-.05in} 85\%$ under varying initial conditions and perception noise through the use of RL in simulation given dense reward engineering and curriculum learning. While IndustReal is a state-of-the-art result in zero-shot transfer of the RL policy trained in simulation, success rate can be further improved, especially for workspace setups that are less carefully instrumented. 

\begin{figure}[t]
    \centering 
    \includegraphics[width=0.5\textwidth]{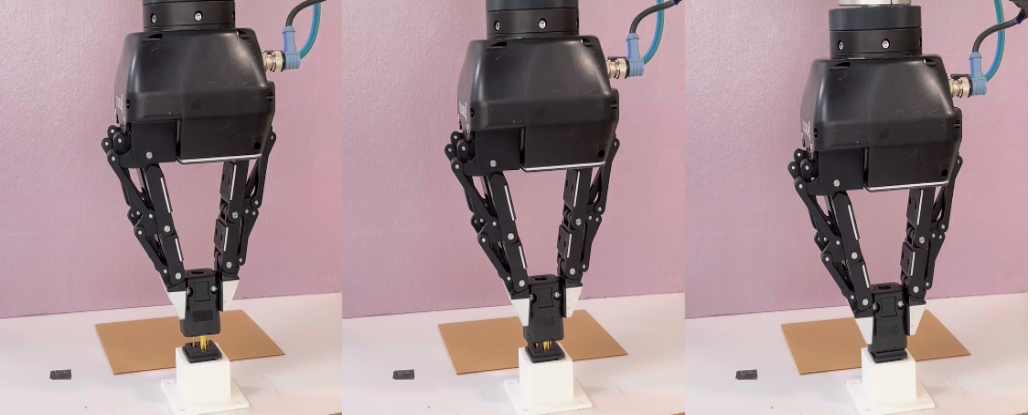}
    \vspace{-.2in}
    \caption{\small Zero-shot transfer of the policy learned in simulation to a real forceful insertion of an unseen plug and socket.}
    \vspace{-.25in}
    \label{fig:robot_setup}
\end{figure}

The {\bf current work} seeks to achieve higher success rates in tight insertion tasks while minimizing engineering effort and enabling zero-shot transfer from simulation to reality. This is accomplished through the integration of model-based reasoning, RL and other key components:

\noindent 1. The method begins with a straightforward potential field policy that operates over the {\tt SE(3)} pose observations of the plug and socket. This \emph{model-based policy} evaluates to near-perfect success rates under zero observation noise in simulation, but its success rate decreases drastically with observation noise. 

\noindent 2. A \emph{residual policy} is trained using a reinforcement learning (RL) objective with only sparse, goal-reaching rewards. The output of the residual policy is added to  the output of the model-based policy. This residual policy is trained in a physics-based simulator (IsaacGym \cite{makoviychuk2021isaac}) with noisy pose observations. The policy is trained to correct for errors introduced by perception noise and unforeseen contacts during real-world execution. A \emph{curriculum-based training scheme} incrementally increases the difficulty of the task, adjusting the noise level of the plug's pose observations and the magnitude of the residual RL policy’s actions. As noise increases, the policy relies more on the RL component and less on the model-based potential field, ultimately balancing out their contributions.

\noindent 3. The combined policy is then transferred directly from simulation to a real robot without fine-tuning. A vision-based pose estimation module detects the socket’s configuration and tracks the plug’s pose. The {\tt SE(3)} plug transforms are returned by the policy and converted into joint controls for the robotic arm controller.

The accompanying experiments demonstrate that this integrated approach significantly improves insertion task success rates compared to alternatives, including IndustReal \cite{tang2023industreal}.

\section{Related Work}
\label{sec:Related_Work}


This section reviews prior efforts on tight insertion, ranging from  model-based to RL-based techniques.

\textbf{Model-based Insertion Strategies} Classical approaches for robotic tight insertion rely on model-based planning and control, that vary from integrating manipulation primitives for fine assembly \cite{suarez2016framework} to assembly-by-disassembly \cite{tian2022assemble} and continuous visual servoing \cite{haugaard2021fast}. Continuous object tracking has also been integrated with passively adaptive mechanical hardware for tight insertion \cite{morgan2021vision}. In general, active and passive compliance can be beneficial for insertion \cite{wang2019robotic,liu2021compliant, morgan2023towards}. Some efforts focus on contact-based search strategies, such as spiral and random motions. Various frameworks have been proposed to discover such solutions, including Finite-State Machine Controllers \cite{wang2023pomdp}, Task-and-Motion Planning \cite{chen2020integrating} and tactile-based behavior trees~\cite{Wu2024Tree}. Search assembly strategies have been evaluated given position uncertainty estimation~\cite{chhatpar2001search}. Socket-location probability distributions can be estimated to devise a search trajectory~\cite{kang2022uncertainty}. Deploying the aforementioned model-based strategies for insertion in the real world can be challenging due to pose uncertainties of the plug and the socket.


\textbf{Machine Learning Insertion Strategies} Data-driven controllers can help address the above challenges. Various methods which utilize multi-modal sensory input \cite{spector2021insertionnet, Spector2021InsertionNetA, multimodal}, learn robust insertion policies from human demonstrations \cite{wen2022you, nair2023learning} and generalize over object geometries \cite{tang2024automate}. Large-scale, high-fidelity simulation \cite{tang2023industreal} can capture the wide distribution of contacts that may be encountered in the real world. The learning process can be accelerated when a well-defined curriculum is used for the RL training \cite{beltran2022accelerating}. A related effort learns motion primitives for insertion \cite{zhang2021learning}. Contact-rich data can be exploited by training with tactile stimuli \cite{dong2021tactilerl}, force/torque measurements \cite{azulay2022haptic}, or a representation of extrinsic contacts like Neural Contact Fields (NCF) \cite{Higuera2023PerceivingEC}.

\textbf{Integrated Control and ML Insertion Strategies} The aforementioned learning-based techniques, however, exhibit high data requirements, especially for tight tolerances. To improve sample efficiency, prior work integrates RL and classical control by using impedance controllers for assembly tasks \cite{Kulkarni2022AIS,khader2020stability}. Using RL to learn a residual policy given a model-based policy can improve sample complexity\cite{Johannink2018ResidualRL, natural}. Furthermore, these methods can work with demonstration data, dynamic movement primitives \cite{Davchev2020ResidualLF, 10000148} and contact-aware, compliant feedback-based controllers \cite{brahmbhatt2023zeroshottransferhapticsbasedobject}. A key advantage of the proposed approach is that it trains entirely in simulation using a simple model-based policy, achieving a high success rate in real-world tight tolerance insertion tasks.

\begin{figure*}[tbp]
\centering
\includegraphics[width=\linewidth, height=0.4\linewidth]{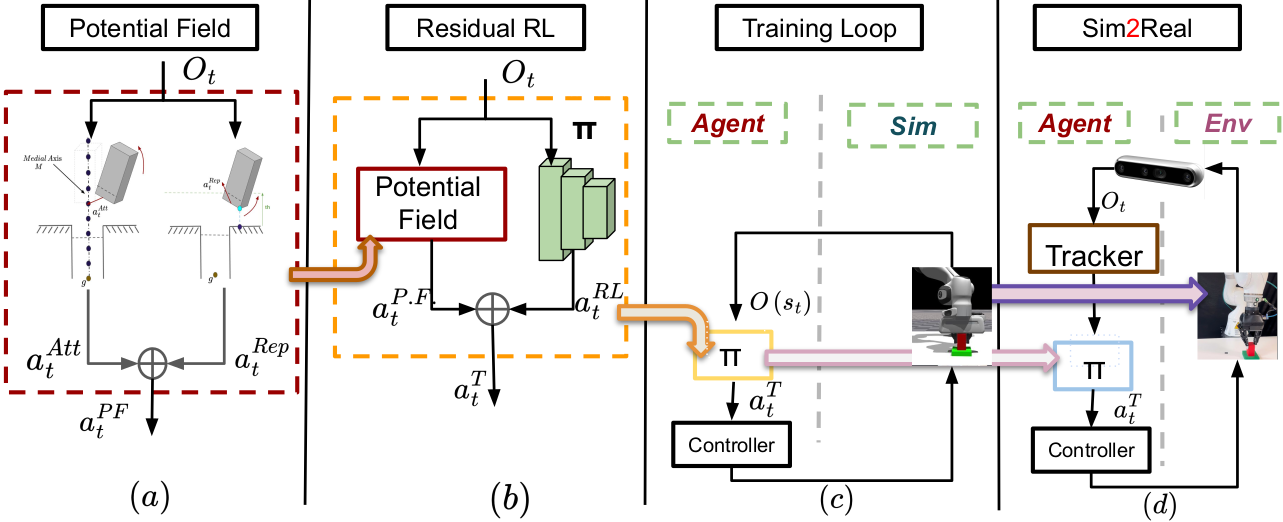}
\caption{\small From left to right: (a) A model-based policy is defined that generates a vector field under full observability. (b-c) An RL policy is trained in simulation given noisy pose observations to provide a residual action that is added to the output of the model-based policy. A sparse reward is provided only upon successful insertion (d) The final policy $\pi$ is zero-shot transferred to the real world, where observations come from a pose tracking module given RGB-D data. A controller translates the policy into robot joint controls.}
\label{fig:overview}
\vspace{-.2in}
\end{figure*}

\section{Method}
\label{sec:Method}

The robot is tasked to insert a grasped object (plug) into a receptacle (socket) with sub-mm tolerance placed firmly in the workspace. At every timestep $t \in [0, T]$, the state is defined as $s_t^{P}$, where $s_t^{P} \in {\tt SE(3)}$ is the plug's pose at timestep $t$. Given the socket's (static) pose $s^{S} \in {\tt SE(3)}$, the goal pose  for the plug so that it is fully inserted into the socket is denoted by $s^{P}_G$. The available observations $o_t = \{ o_t^{P}, o^{S}\}$ correspond to continuous estimates $o_t^{P} \in {\tt SE(3)}$ of the plug's pose and an estimate $o^{S} \in {\tt SE(3)}$ of the socket's pose given visual input. The 3D object models of both the plug and socket are known at the time of execution and are denoted by $\Gamma^{P}$ and $\Gamma^{S}$ respectively. The objective is to train a policy $\pi(o_t)$, which, during inference, given an observation $o_t$, outputs an action $a \in {\tt SE(3)}$  that corresponds to transformations of the plug's pose so that it eventually reaches $s^{P}_G$. 

Fig.~\ref{fig:overview} outlines the components of the proposed approach for computing policy $\pi(o_t)$: (i) a simple model-based policy outputs an action $a_t^\text{PF}$ using a potential field -- this is computed at every timestep during training and inference in simulation and reality for the target geometries $\Gamma^{P}$ and $\Gamma^{S}$; (ii) a residual RL policy's action $a_t^\text{RL}$ is added to the model-based policy's action to provide the final output action $a_t^T$; (iii) training is performed in simulation over randomized conditions to learn $a_t^\text{RL}$ so that $a_t^T$ results in successful insertions given sparse rewards and a curriculum; and, finally, (iv) the resulting policy $\pi(o_t)$ is transferred to the real system to solve tight insertion tasks involving  novel geometries relative to those seen during training. 




\subsection{Model-based Policy} 

The potential-field action $a_{t}^\text{PF}$ is computed given the plug and socket's observed poses $(o_t^{P}, o^{S})$ and their geometries ($\Gamma^{P}$, $\Gamma^{S}$). $a_{t}^\text{PF}$ is a combination of an action arising from an attractive potential $a_{t}^\text{Att}$, i.e., moving the plug towards the goal $s^{P}_G$, and an action arising from a repulsive potential $a_{t}^\text{Rep}$, i.e., pushing the plug away from collisions with the socket. 

\begin{figure}[h]
\vspace{-0.1in}
    \centering
    \includegraphics[width=0.4\textwidth]{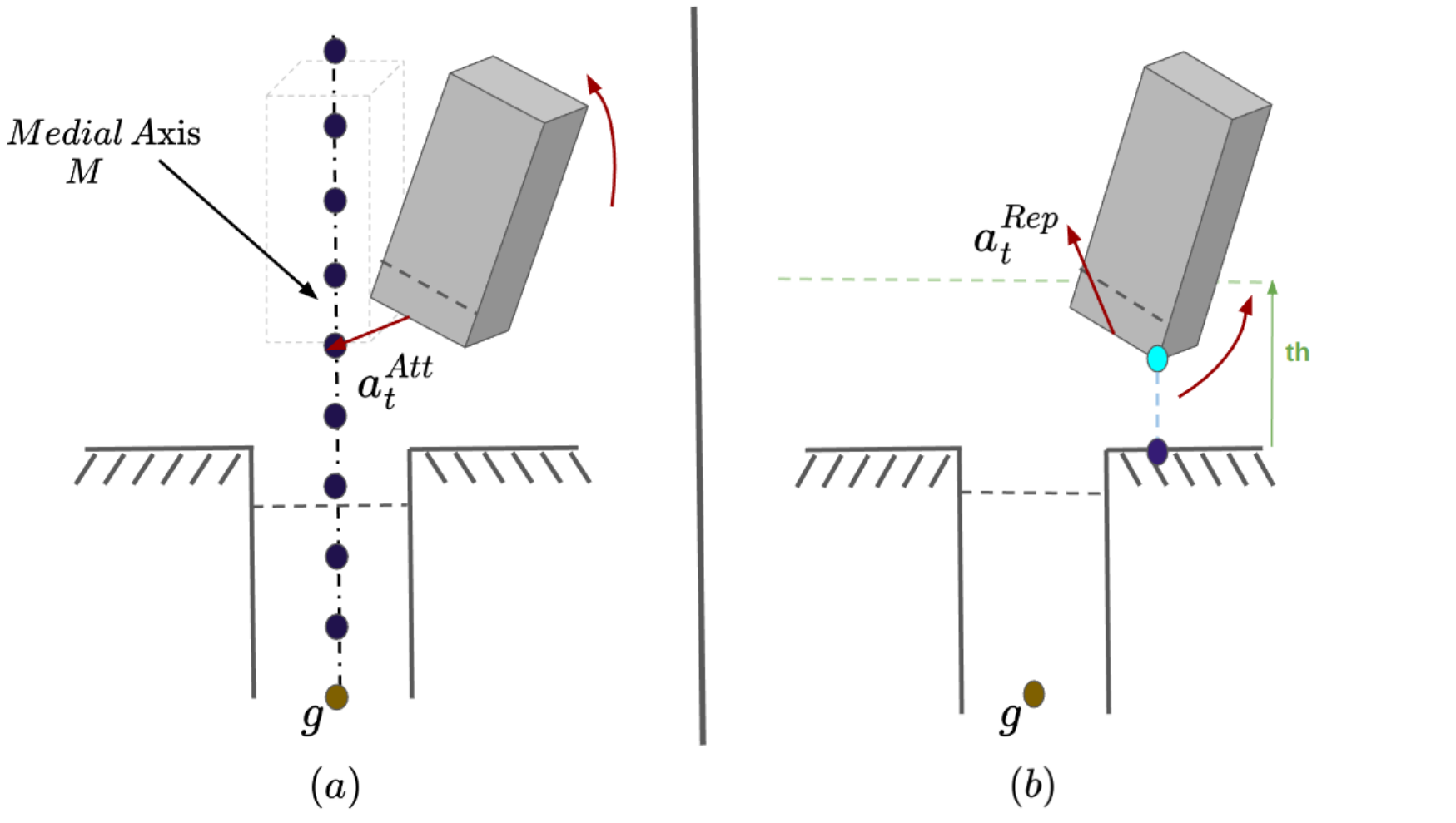}
    \vspace{-.1in}
    \caption{(left) The attractive potential field moves the object towards a nominal, straight-line insertion trajectory that leads to the goal pose; (right) the repulsive component pushes the object away from making contact with the socket, only when the object is close to it.}
    \label{Fig:PF_Scetch}
    \vspace{-.15in}
\end{figure}

\textbf{Attractive Component} A nominal, collision-free path for the plug is defined to connect the goal plug pose $s^{P}_G$ to a pose with the same orientation above the socket along a straight retraction path, as in Fig. \ref{Fig:PF_Scetch}. This nominal retraction path is along the socket's medial axis, i.e., the locus of equidistant points from the socket's inner walls. Then, $k$ \textit{anchor} poses are defined along the nominal path by discretizing it. For every possible observation of the plug's pose $o_t^{P}$, the attractive potential computes the closest anchor pose on the nominal path $s_{cl}$. If the distance between $o_t^{P}$ and $s_{cl}$ is above a threshold, then the attractive potential returns $a_{t}^{\text{Att}} = s_{cl} - o_t^{P}$. If the distance $s_{cl}$ is below a threshold, then the anchor pose $s_{next}$ along the nominal path that is closer to the goal than $s_{cl}$ is selected as the target. In this case, the attractive potential returns an action vector $a_{t}^{\text{Att}} = s_{next} - o_t^{P}$. Thus, the attractive field points towards the nominal path far from it and points more towards the goal pose close to the nominal path.



\textbf{Repulsive Component} The closest pair of points on the plug $p_t^P$ and the socket $p_t^S$ are identified. If their distance $d_{t} = ||p_t^P - p_t^S||$ falls below a threshold $th$, a repulsive action is applied at the plug geometry's origin (which here is defined to be the plug's bottom-center point), to move the plug away from the socket. Otherwise, i.e., when $d_{t} > th$, the repulsive component is zero. To compute the repulsive action, a virtual 3D force vector $v_t = p_t^P -p_t^S $ is computed. A distance-normalized version of the virtual force vector $v_t$ is defined as $\mathcal{N}(v_t)$ and corresponds to a division of the vector's magnitude with the distance $d_{t}$. This has the effect that the magnitude of $\mathcal{N}(v_t)$ increases as the peg approaches the socket. Then, the repulsive action is computed as $a_{t}^{\text{Rep}} = \mathbf{J} \cdot \mathcal{N}(v_t)$, where $\mathbf{J}$ is the Jacobian matrix that relates the coordinates of $p_t^P$ to the plug's frame. This component moves the plug away from contact states that prohibit task success in tight setups. Overall, its use reduces the need to carefully tune the hyperparameters of the Attractive Field.

\textbf{Potential Field} The overall action combines the attractive and repulsive actions with a weighted sum, where weights $w^{Tr.}$ and $w^{Rot.}$ $\in$ [0,1] (same for all objects) are applied to the translational and rotational components.

\subsection{Residual RL}
\label{sec:resRL}

\begin{figure}[t]
    \centering
    \includegraphics[width=\linewidth]{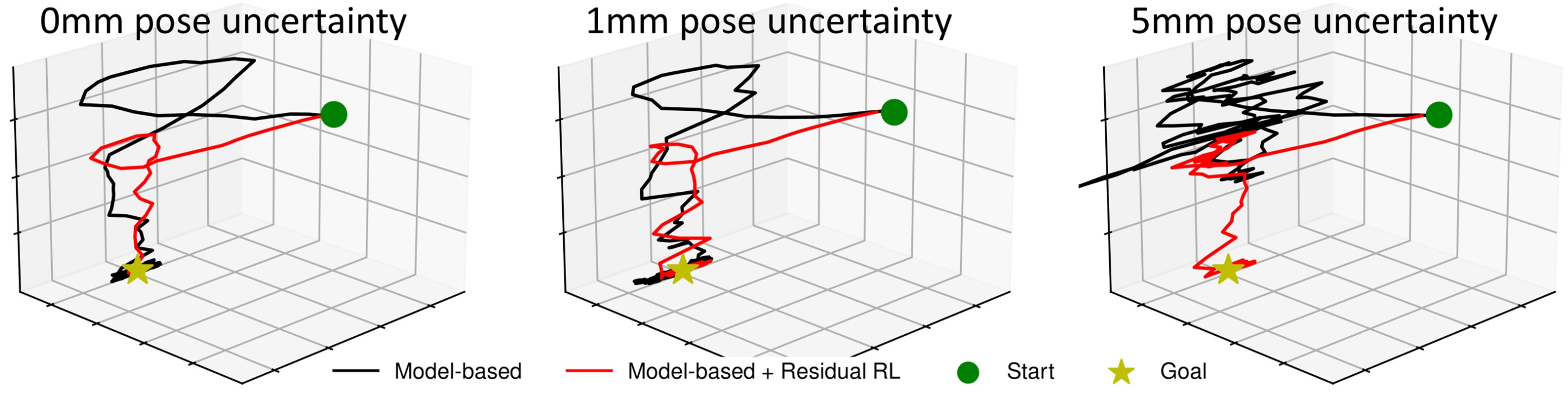} 
    \caption{\small Impact of observation noise on insertion trajectories: (left) Under no noise, both the model-based controller with and without the residual policy succeed. Residual RL helps to shorten trajectories.  (center) With low observation noise, the performance of the model-based controller declines, but in combination with the residual policy output, the performance is preserved. (right) At high levels of noise, the model-based controller fails, while integrating the residual RL policy effectively compensates for the noisy pose estimate.}
    \label{fig:Effect_Obs_Noise_PF}
    \vspace{-.3in}
\end{figure}

The potential field actions succeed in insertion when the ground-truth poses of the plug and socket are available, e.g., in simulation. When these pose estimates are noisy, as in the real world, the efficacy of the potential field-based policy declines drastically (see Fig.\ref{fig:Effect_Obs_Noise_PF}). To address this, complementary actions $a_{t}^{RL} \in {\tt SE(3)}$ are generated by a residual Deep RL policy and added to the potential field action. The RL policy accounts for uncertain estimations, enabling successful task completion. The combined action $a_{t}^{T} = a_{t}^{PF} + \beta  a_{t}^{RL}, \textnormal{where } \beta \in [0,1]$ scales the contribution of the two action components.

\textbf{Sparse Rewards} The model-based policy enables the use of a sparse goal-reaching reward for training the residual Deep RL component. A fixed positive reward is provided if the plug is fully inserted into the socket. In addition, a negative reward is defined for object-object inter-penetration during contact as IssacGym allows for significant inter-penetration between objects as noted in IndustReal \cite{tang2023industreal}. This discourages RL from exploiting the simulation during training.

\textbf{Scaling and Noise Curriculum Training} While training the RL policy in simulation, uniform noise is added to the ground-truth poses of both the plug and the socket. A curriculum training strategy is implemented, where the plug observation noise ranges from 0mm/0$^\circ$ to $n_{max}$mm/$n_{max}^\circ$, while the scaling parameter $\beta$ ranges from $0$ to $1$. This curriculum adapts the difficulty of the training task by observing the success rate of insertion across multiple trials.  It increases the difficulty if the success rate exceeds 75$\%$ and decreases it if the success rate falls below 50$\%$. The noise ranges increment or decrement by $st$ respectively. This adaptive approach trains the Deep RL policy to increase its contribution towards the combined action as observation uncertainty increases. During inference, the scaling parameter $\beta$ is set to 1.

\textbf{RL architecture } An asymmetric actor-critic architecture is employed~\cite{pinto2017asymmetric}. The actor network consists of a 3-layer Multi-Layer Perceptron (MLP) and a 2-layer Long Short-Term Memory (LSTM). The critic network consists of a 3-layer MLP. Both networks input the estimated plug and socket poses, and the critic additionally receives their ground-truth poses as privileged information. The architecture is trained with Proximal Policy Optimization (PPO) \cite{schulman2017proximal}. 


\textbf{Training Randomization and Noise Conditions} During training, the socket pose is randomized within a range of $\pm 10$ cm in the x-y plane, 5 cm in the z-axis, and $\pm 5^\circ$ in yaw. The initial plug pose varies within $\pm 10$ mm in the x-y plane and $\pm 15^\circ$ across roll, yaw, and pitch, while being positioned 10 mm above the socket's tip. To simulate uncertain observations, the 6D observation noise for both the plug and the socket is sampled i.i.d at each time-step from a uniform distribution. This noise is constrained to a maximum of $\pm 5$ mm/$5^\circ$ for the plug and $\pm 1$ mm/$1^\circ$ for the socket. The noise curriculum step-size is set to $st=$ 0.1mm.

\subsection{Sim2Real Transfer and Real-World Components}
\label{sec:Implementation}

The policy was trained using IsaacGym \cite{makoviychuk2021isaac} on a model of the Franka Emika Panda robot with a task impedance controller. The policy was then deployed zero-shot in the real world on a Kuka iiwa 14 manipulator with a Robotiq 3-fingered gripper, that uses a position controller. The successful transfer between the disparate simulation and real-world setup is facilitated by the definition of actions over the plug's {\tt SE(3)} pose space.



\textbf{RGB-D Pose Tracker and Pose Control} M3T~\cite{stoiber2022multi}, an RGB-D-based pose tracker, provides an estimate of the socket's pose at the beginning of each trial as well as dynamically tracking the plug's pose across the trial (at a frequency of 30Hz). Tracking accuracy reduces as the plug nears and engages with the socket, due to increased occlusions. The combination of the deployed tracker with the proposed approach, results in a high insertion ratio. Task failures due to object-gripper slippage \cite{tang2023industreal} are addressed as the policy reasons about the {\tt SE(3)} pose of the plug and socket in a closed-loop manner.

\begin{figure}[ht]
    \vspace{-.1in}
    \centering
    \includegraphics[width=\linewidth]{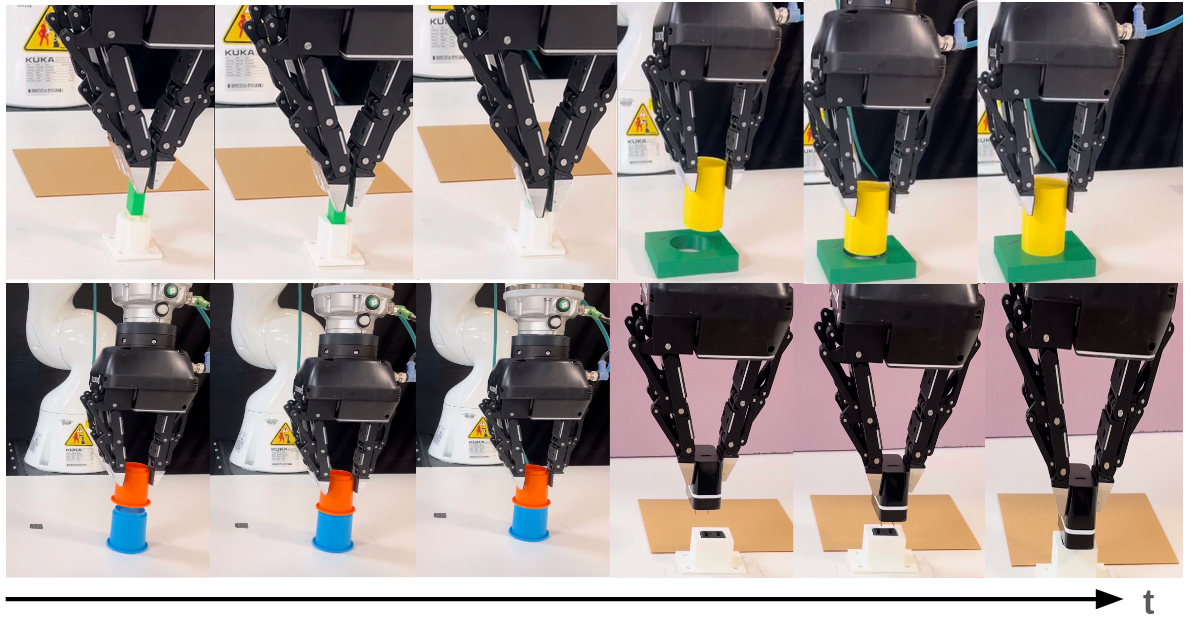}
    \vspace{-.25in}
    \caption{\small Sim2Real policy transfer with 2 known (top) and 2 unknown at training objects (bottom).}
    \label{fig:unique_label} 
    \vspace{-.15in}
\end{figure}

\section{Results}
\label{sec:result}

\begin{figure}[h]
    \centering
    \begin{minipage}[b]{0.17\textwidth}
        \centering
        \includegraphics[height=0.42in]{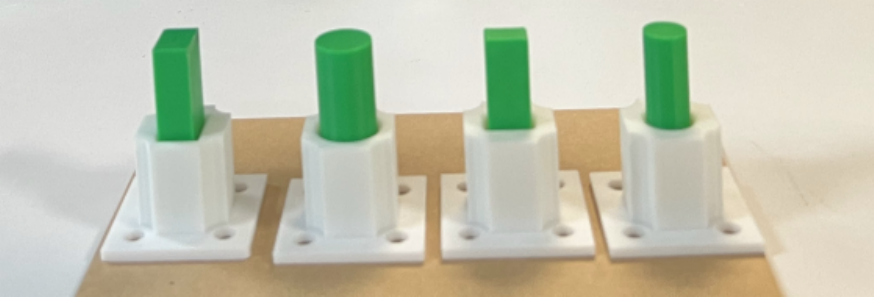}
    \end{minipage}%
    \begin{minipage}[b]{0.17\textwidth}
        \centering
        \includegraphics[height=0.42in]{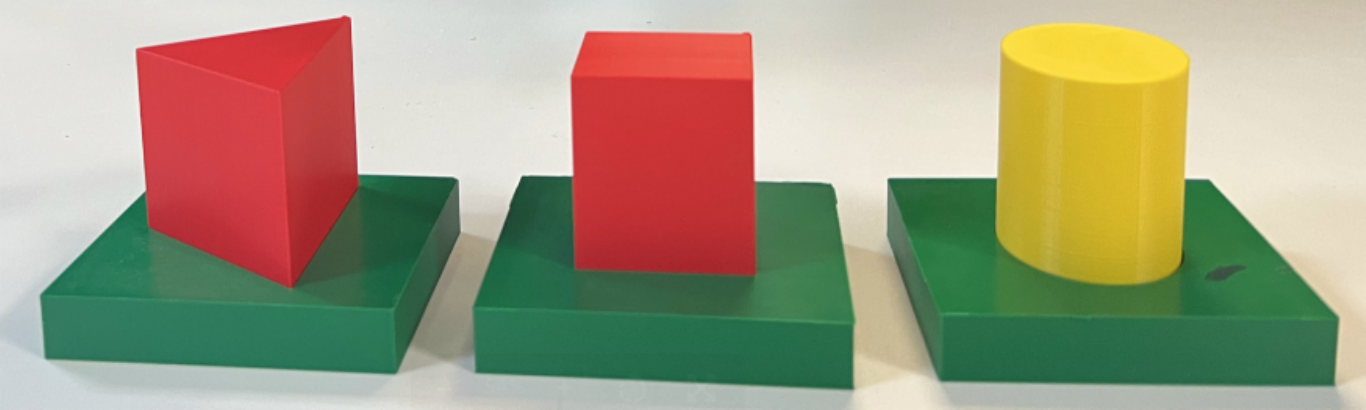}
    \end{minipage}%
    \begin{minipage}[b]{0.17\textwidth}
        \centering
        \includegraphics[width=0.8in, height=0.42in]{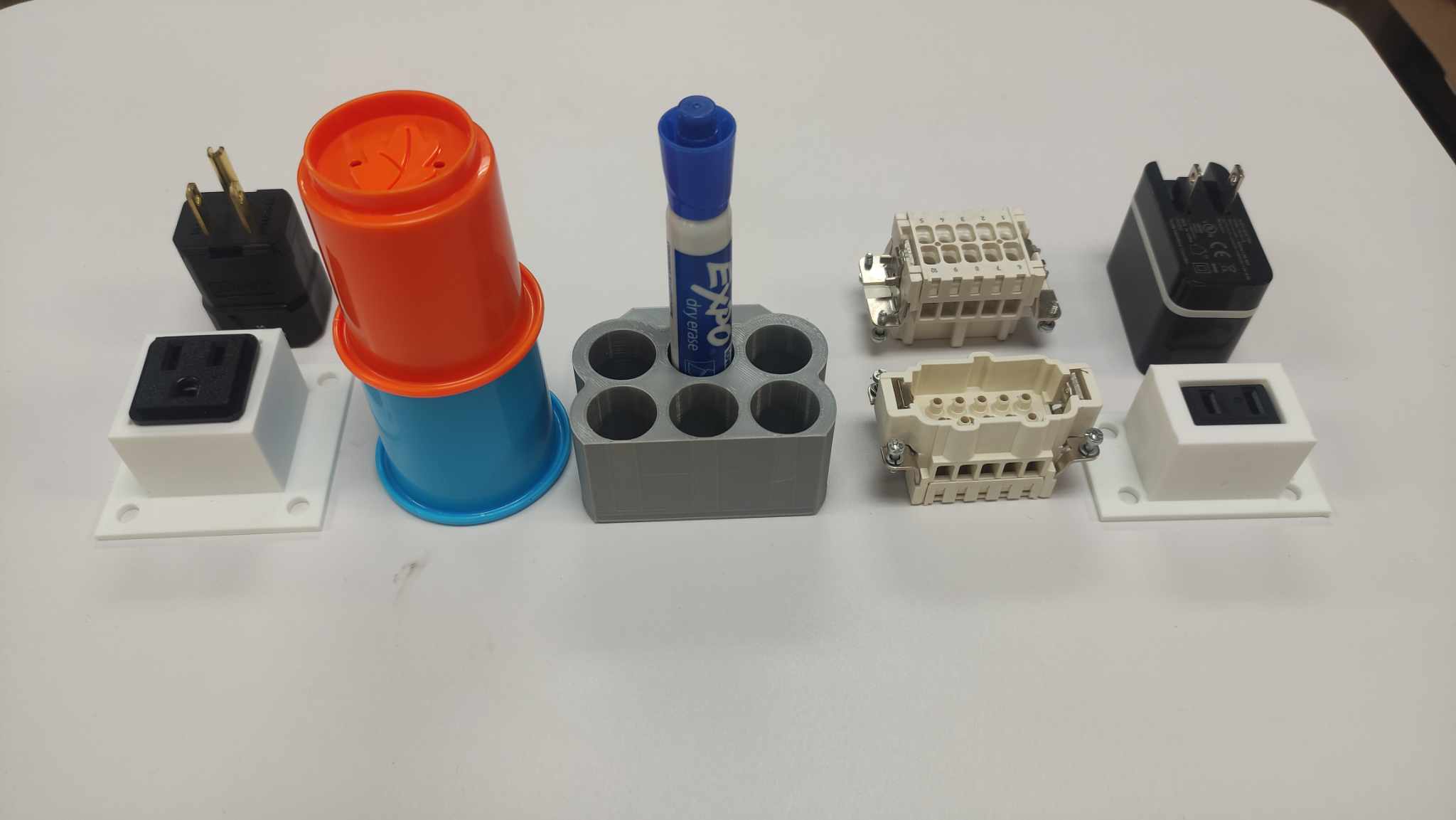}
    \end{minipage}
    \caption{\small (left) 3D printed objects from IndustReal \cite{tang2023industreal} with $0.5-0.6$ mm tolerance. (middle) 3D printed custom objects with 2 mm (Easy), 1 mm (Medium), and 0.1 mm (Hard) tolerance. (right) Household objects not seen during training.}
    \vspace{-.1in}
    \label{fig:objects}
\end{figure}

\textbf{Evaluated Object Categories} The evaluation is performed across three plug-socket categories. The first category (Fig.\ref{fig:objects} (left)) includes small cylindrical and rectangular plugs of widths 8, 12, and 16 mm with tolerances of approximately 0.5$-$0.6 mm, similar to NIST Taskboard Challenge benchmark \cite{kimble2020benchmarking} used for evaluation by IndustReal \cite{tang2023industreal}. The second category (Fig.\ref{fig:objects} (middle)) encompasses larger cylindrical and rectangular sockets of$~$50mm width categorized into three difficulty levels based on their tolerance: Easy ($\sim$2 mm), Medium ($\sim$1 mm), and Hard ($\sim$0.1 mm). The last category of objects (Fig.\ref{fig:objects} (right)) is only used for real-world trials, and corresponds to five household objects that have not been seen during training: a 2-prong charger, a 3-prong charger, a HAN-type connector, two types of cups, and a marker with a marker holder.

\begin{table*}[t] 
\caption{Insertion Success Rates in Simulation \vspace{-.05in}}
\label{tab:Simulation_Comparison}
\centering
\scriptsize 
\setlength{\tabcolsep}{6pt} 
\begin{tabular}{@{}p{3.8cm}p{1.8cm}p{1.8cm}p{1.8cm}p{1.8cm}p{1.8cm}p{1.8cm}@{}}
\toprule
& \multicolumn{3}{c}{Fig. \ref{fig:objects} (left) Objects} & \multicolumn{3}{c}{Fig. \ref{fig:objects} (middle) Objects} \\
\cmidrule(lr){2-4} \cmidrule(lr){5-7}
Method & \small{0mm}/$0^{o}$ & \small{1mm}/$1^{o}$ & \small{5mm}/$5^{o}$ & \small{0mm}/$0^{o}$ & \small{1mm}/$1^{o}$ & \small{5mm}/$5^{o}$ \\
\midrule
\small{IndustReal \cite{tang2023industreal}} & 92.40$\pm$2.30\% & 88.60$\pm$2.41\% & n.a. & 26.77$\pm$13.88\% & 27.09$\pm$13.61\% & n.a. \\
\small{PF + Res.RL + Curr. of \cite{Kulkarni2022AIS}} & 98.65$\pm$0.87\% & \textbf{98.44$\pm$0.55\%} & \textbf{97.50$\pm$0.65\%} &  95.28$\pm$3.39\% & 92.36$\pm$4.35\% & 33.87$\pm$6.66\% \\
\small{\textbf{Ours}} & \textbf{100.0$\pm$0.0}\% & 96.10$\pm$1.92\% & 96.25$\pm$1.22\% & \textbf{99.09$\pm$0.91\%} & \textbf{98.28$\pm$0.65\%} & \textbf{95.88$\pm$1.86\%} \\
\bottomrule
\end{tabular}
\vspace{-.2in}
\end{table*}

{\bf Evaluation in IsaacGym} Table~\ref{tab:Simulation_Comparison} evaluates the proposed method and the alternatives in simulation. The evaluation metric is the percentage of successful insertions for different levels of maximum perception noise $n_\text{max}$. Noise is sampled uniformly around the true states up to the value $n_\text{max}$. Three different values of $n_\text{max}$ are considered for plug translational/rotational noise respectively: 0mm/$\ang{0}$, 1mm/$\ang{1}$, and 5mm/$\ang{5}$. For all scenarios with non-zero plug observation noise, a corresponding noise of 1mm/$\ang{1}$ was added to the socket. 

The proposed method is compared against IndustReal~\cite{tang2023industreal}, which also zero-shot transfers from Sim2Real but is an exclusively RL approach. The code for AutoMate \cite{tang2024automate}, an extension of IndustReal was not available while carrying out this evaluation. The proposed approach is also evaluated when using an alternative curriculum that transitions from model-based control to residual RL over time, instead of as a function of noise \cite{Kulkarni2022AIS}. 

While evaluating IndustReal, observation noise of 1mm/$\ang{1}$ is added only to the socket pose, as IndustReal operates over the {\tt SE(3)} pose of the end-effector, whereas the proposed method  operates over the {\tt SE(3)} pose of the plug and socket. Thus, for IndustReal, the results reported with 1mm/$\ang{1}$ noise are taken directly from the publication. IndustReal could not be evaluated with 5mm/$\ang{5}$ plug noise scenario as applying high noise to the socket's pose artificially collapses its performance.

For the objects in Fig. 6 (left), a single policy was trained across all objects for a fair comparison with IndustReal. For the objects in Fig. 6 (middle), however, a dedicated policy was trained for each object instance to prevent the easier geometries from
inflating the success rates while inserting the more challenging objects. The evaluation task is to insert these objects with the smallest tolerance. Similar to IndustReal, all policies are trained and tested over 5 random seeds and the mean and standard deviation of the insertion successes are reported. 

IndustReal does not rely on 3D models of the plug and socket, whereas the proposed method requires them. To ensure a fair comparison, the plug and socket are approximated by their largest common bounding shape primitive (box, cylinder, etc.) This approximation allows the proposed method to operate without relying on specific instance 3D models while inserting the objects in Fig.\ref{fig:objects} (left), (Table~\ref{tab:Simulation_Comparison} - left). The Potential Field (PF) policy generates the same actions across all geometric instances, using a 3D bounding box that approximates actual models. Since IndustReal trains a single policy across all these objects, this adjustment ensures alignment in model requirements for a fair comparison. Given this setup, the residual RL component should also compensate for the lack of a known 3D object model. This approximation applies only to the PF controller, in simulation, where no pose estimation is required. During Sim2Real transfer, the trackeruses the full 3D object models.

Overall, the proposed method consistently outperforms IndustReal in simulation. The time-based curriculum \cite{Kulkarni2022AIS} also achieves high insertion success percentages for the objects of Fig.\ref{fig:objects}(left), which verifies the efficacy of the designed model-based controller. The proposed success-based curriculum strategy surpasses the time-based curriculum for all for the objects of Fig.~\ref{fig:objects}(middle), and a single object of Fig.~\ref{fig:objects} (left). As the difficulty of the task increases, it is observed that the difference in performance between the proposed method and the comparison points becomes more evident. 

Note that the proposed residual RL policy requires only 25$\%$ to 33$\%$ of the training time/samples reported in the IndustReal work, without access to a handcrafted dense reward function. Our policy was fully trained in 2-3 hours using a single GPU, while IndustReal reports a corresponding time of 8-10 hours.

\begin{table*}[t] 
\caption{Ablation Study}
\label{tab:Ablation}
\centering
\scriptsize 
\setlength{\tabcolsep}{6pt} 
\begin{tabular}{@{}p{3.8cm}p{1.8cm}p{1.8cm}p{1.8cm}p{1.8cm}p{1.8cm}p{1.8cm}@{}}
\toprule
& \multicolumn{3}{c}{Fig. \ref{fig:objects} (left) Objects} & \multicolumn{3}{c}{Fig. \ref{fig:objects} (middle) Objects} \\
\cmidrule(lr){2-4} \cmidrule(lr){5-7}
Method & \small{0mm}/0° & \small{1mm}/1° & \small{5mm}/5° & \small{0mm}/0° & \small{1mm}/1° & \small{5mm}/5° \\
\midrule
PF & 98.91$\pm$0.89\% & 99.84$\pm$0.35\% & 3.28$\pm$1.69\% & 97.55$\pm$1.63\% & 97.81$\pm$1.58\% & 46.51$\pm$4.28\% \\
PF + Learned Scaling w & 90.47$\pm$3.51\% & 94.69$\pm$2.02\% & 11.72$\pm$2.45 \% & 89.21$\pm$2.44\% & 90.60$\pm$3.28\% & 7.76$\pm$3.71\%\\
PF + Res.RL & 99.53$\pm$0.70\% & \textbf{100.0$\pm$0.00}\% & \textbf{97.66$\pm$1.56\%} & 82.37$\pm$4.48\% & 63.48$\pm$5.52\% & 61.09$\pm$45.18\% \\
PF + Res.RL + Learned Scaling $\beta$ & 98.28$\pm$0.65\% & 82.03$\pm$2.27\% & 84.69$\pm$4.12\% & 93.90$\pm$1.50\% & 92.34$\pm$2.67\% & 70.15$\pm$2.23\% \\
\textbf{Ours} & \textbf{100.0$\pm$0.00}\% & 96.10$\pm$1.92\% & 96.25$\pm$1.22\% & \textbf{99.09$\pm$0.91\%} & \textbf{98.28$\pm$0.65\%} & \textbf{95.88$\pm$1.86\%} \\
\bottomrule
\end{tabular}
\vspace{-.1in}
\end{table*}

\begin{figure}[h]
    \centering
\vspace{-.05in}    
\includegraphics[width=0.45\textwidth, keepaspectratio]{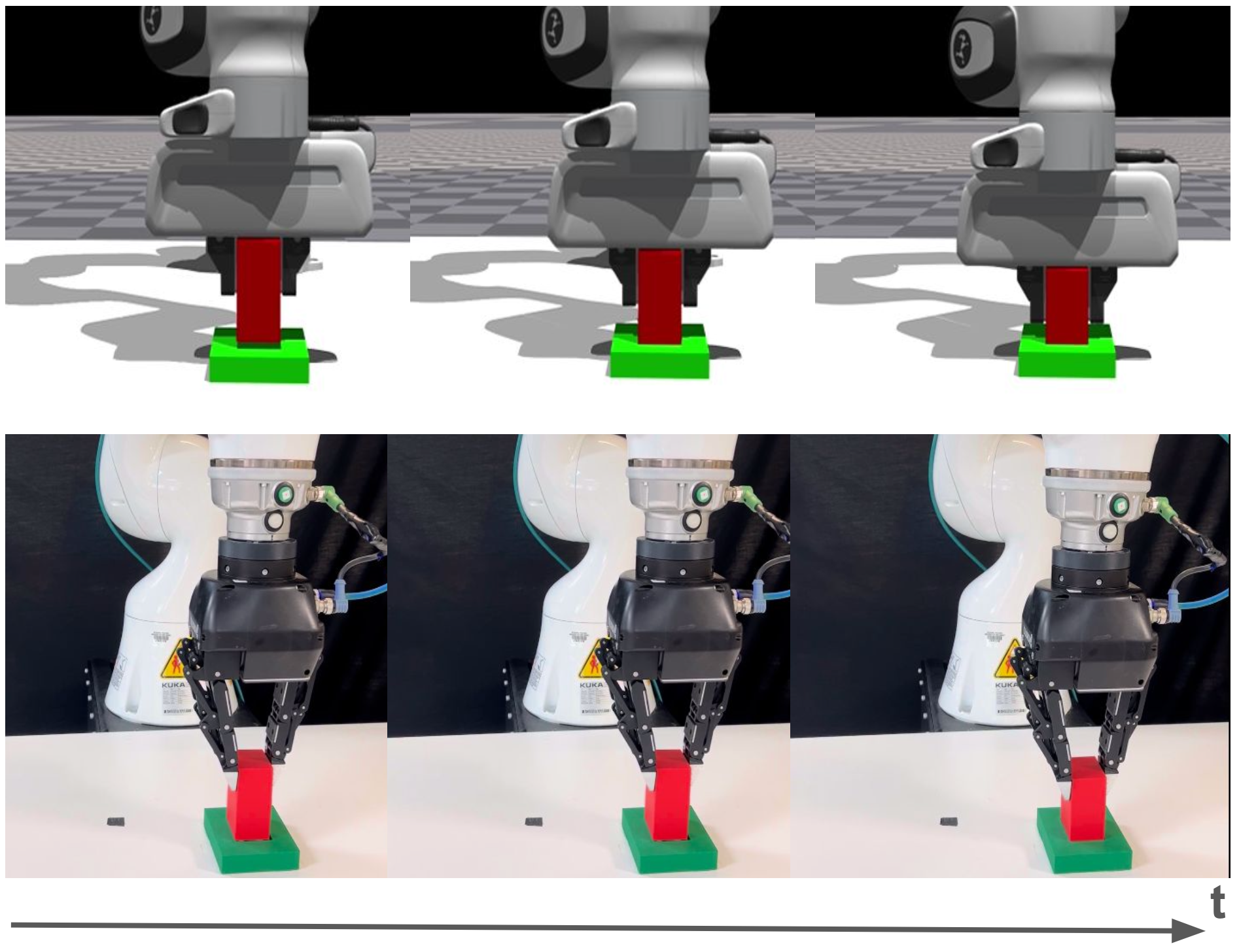}
    \caption{\small Plug insertion in simulation (top) and real world (bottom).}
    \label{fig:rect_sim2real}
    \vspace{-.1in}
\end{figure}


\noindent {\bf Ablation over Residual RL strategies:} Table \ref{tab:Ablation} evaluates variations in implementation of the proposed approach with progressively increasing $n_\text{max}$ values to gauge the impact of noise levels. The first variation corresponds to applying only the Potential Field (PF) policy. Then, an alternative approach is tested where RL is used to learn the weights $w^{Tr.}$ and $w^{Rot.}$ that combine the attractive and repulsive components of the PF. Following this, variations of the proposed method are evaluated with and without the proposed success-based curriculum strategy.  All variations of the proposed approach, where residual actions are output by the deep RL module, achieve high insertion rates for the tasks of Fig.\ref{fig:objects} (left). The full proposed method is the most successful while attemtping the harder insertion task of objects in Fig.\ref{fig:objects} (middle). Simultaneously applying the noise-action curriculum ensures that for every $n_\text{max}$ value, the RL succeeds with the true pose as observations before progressing to more difficult task conditions. Combining this with the proposed success-based curriculum accelerates convergence to a insertion policy that achieves higher success rates than non-curriculum-based training schemes.


\noindent \textbf{Real-world Experimental Setup:} Experiments were conducted with the Kuka iiwa14 7-DOF manipulator in a minimally instrumented environment.  An Intel RealSense D435 RGB-D camera was utilized to monitor the scene, capturing the manipulator as it reached for, grasped, and positioned a target object at an randomized initial pose, 10mm above the socket. The control actions from policy then take over until the object is inserted (or time limits exceed). The socket was rigidly affixed to the table with heavy-load mounting tape to immobilize it.


\begin{figure}[h]
    \vspace{-.1in}
    \centering
    \begin{subfigure}
    {0.49\linewidth}
        \centering
        \includegraphics[width=\linewidth ]{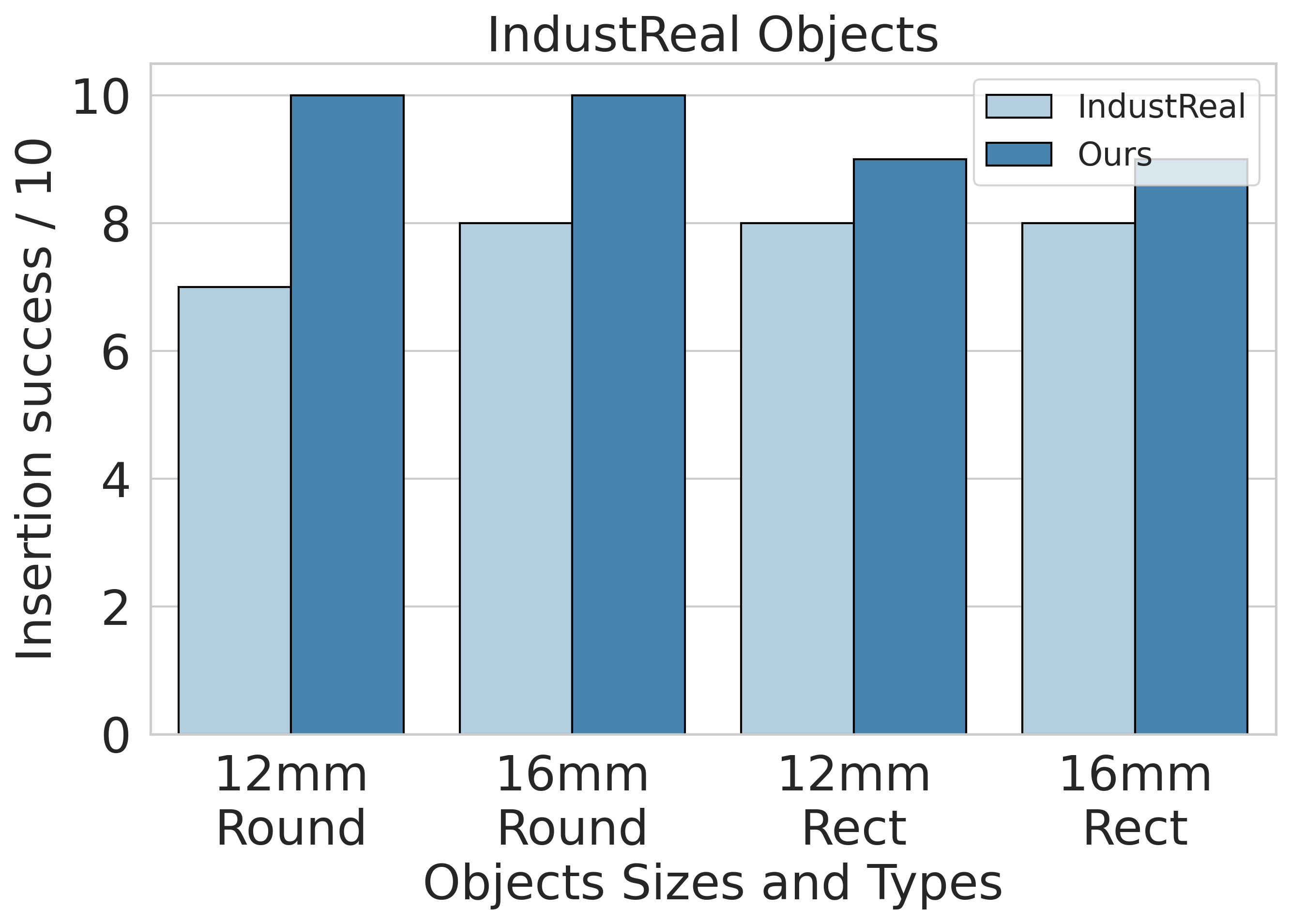}
        \caption{\small Comparison with IndustReal}
        \label{fig:wrapfig1a}
    \end{subfigure}
    \begin{subfigure}{0.49\linewidth}
        \centering
        \includegraphics[width=\linewidth]{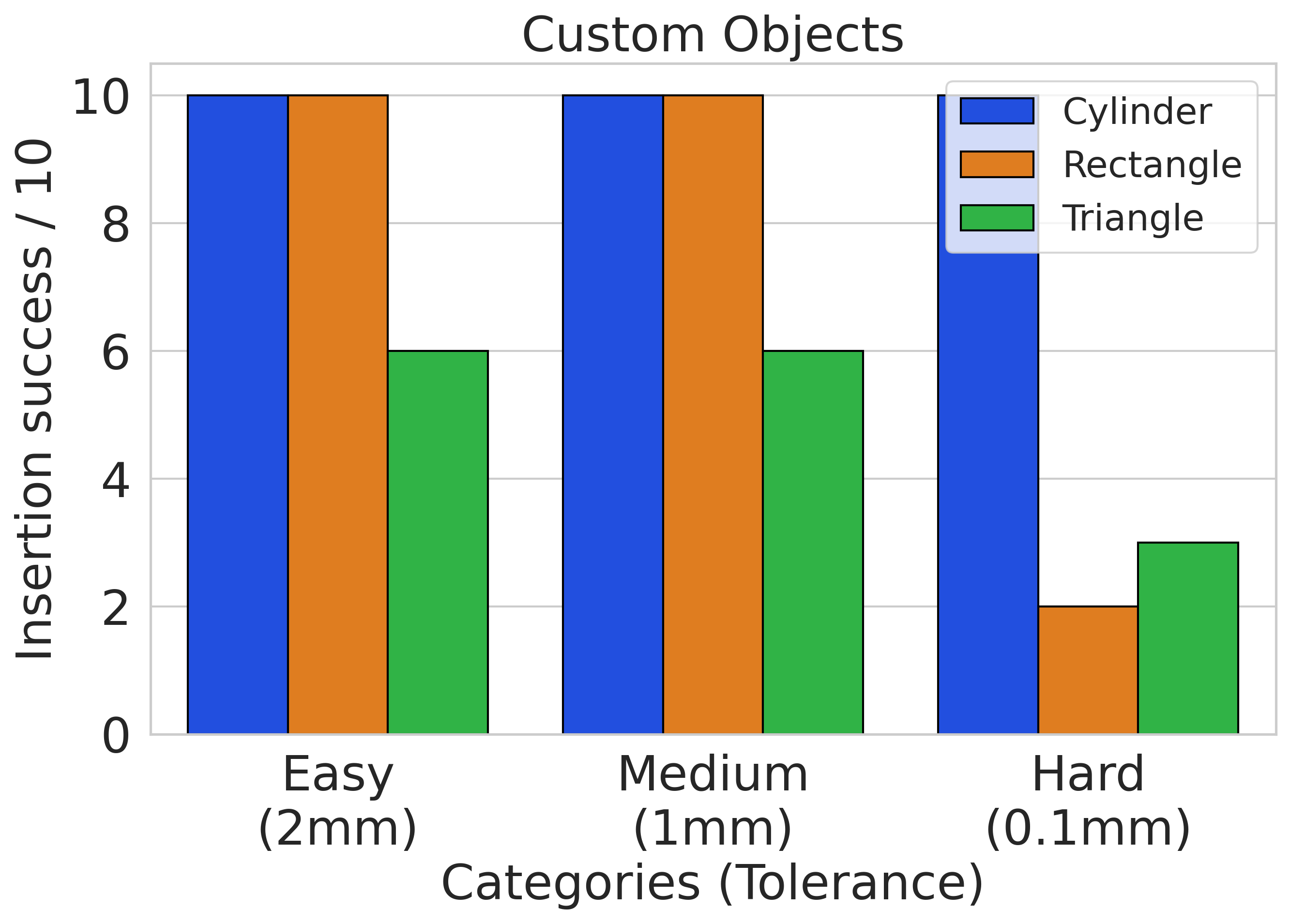}
        \caption{\small Custom objects}
        \label{fig:wrapfig1b}
    \end{subfigure}

    \caption{\small Real-world evaluation: Number of successful insertions over 10 trials for different objects.}
    \label{fig:combined_figures}
    \vspace{-.2in}
\end{figure}

\noindent \textbf{Previously Seen Objects:} Comparison with IndustReal in the real-world is shown in Fig.~\ref{fig:combined_figures} (left). The statistics for IndustReal are derived from the corresponding publication. IndustReal achieves sim-to-real transfer for the same robot model, with a low-level task impedance controller and an instrumented setup. The proposed method achieves higher success rate across objects with less instrumentation and while training over a disparate setup in simulation.

Additional sim-to-real-world trials were performed on the more challenging objects of Fig.~\ref{fig:objects} (middle). The policy was trained only on the Hard objects in simulation and then evaluated in the real world on the Easy, Medium and Hard objects. Fig.~\ref{fig:combined_figures} (right) shows that the proposed policy successfully inserted all instances of the Cylinder effectively (10/10 successful trials). Similar results are shown for the Easy and Medium Rectangular objects, while for the Hard case of $0.1mm$ tolerance, the recorded success rate drops. The Triangular objects were the most challenging with recorded success rates of 6/10, 6/10, 3/10 for the Easy, Medium, and Hardß cases respectively. Rotational alignment of the plugs with respect to the socket is a crucial factor for successful insertions in very low tolerance regimes. 

\noindent \textbf{Unseen household objects:} The proposed method was also tested on 5 unseen real-world objects (Fig.~\ref{fig:objects} right). The policy was trained on the objects of Fig.~\ref{fig:objects} (left) and then tested directly on the household objects, starting from 10 randomized initial plug poses for each object. Fig.~\ref{tab:ood_insertion} shows that the proposed policy consistently achieved high success percentages for all test objects, demonstrating robust performance even in scenarios requiring significant force modulation (e.g., 2-prong, 3-prong, HAN-connector). These results show good generalization to novel geometries. 

\begin{figure}[h]
    \centering
    \begin{tabular}{@{}lcc@{}}
        \toprule
        Object & \textbf{IndustReal} & \textbf{Ours} \\
        \midrule
        2-Prong Charger & 10/10 & 10/10 \\
        3-Prong Charger & 7/10 & 10/10 \\
        \hdashline 
        Cups & - & 10/10 \\
        Marker & - & 10/10 \\
        HAN Connector & - & 9/10 \\
        \bottomrule
    \end{tabular}
    \caption{\small Success rate for unseen household objects. Number of successful insertions over 10 real-world trials each.}
    \label{tab:ood_insertion}
    \vspace{-.2in}
\end{figure}

\section{Discussion}
\label{sec:discussion}

This paper proposes a hybrid approach for robotic insertion, which leverages the strengths of both model-based planning and data-driven methods, using a potential field as a guiding policy that works well in noise-free scenarios. RL enables the system to adapt to noise without requiring complex reward engineering. The policy is trained exclusively in simulation using sparse rewards and transferred zero-shot to the real world with good accuracy.

While in industrial setup 3D models may be available, the reliance on 3D models during inference may limit the method's applicability in service robotics. Future work will explore how to waive this requirement, while still achieving high accuracy and benefiting from model-based reasoning. Furthermore, occlusions prevented tests on small objects, indicating the need for fine-grained sensing. 

Lastly, this work aims to inform how to solve general contact-rich manipulation tasks with tight tolerances, e.g. top-down insertion, while minimizing human engineering.  As IndustReal demonstrated, for top-down insertion it is possible to design useful dense rewards. But it is not obvious how to define dense rewards that work across contact-rich manipulation tasks. The current effort indicates that the combination of a model-based policy and sparse reward residual RL can provide solutions in this domain. For other manipulation tasks, the model-based policy may be defined after a classical motion planner first generates successful (more complex) manipulation solutions under full observability. These model-based policies can still guide residual RL policies that work under partial observability and noise in the real-world.

\clearpage

\bibliographystyle{IEEEtran}
\bibliography{root.bib}

\end{document}